\documentclass{article}

\usepackage{PRIMEarxiv}

\usepackage[utf8]{inputenc} 
\usepackage[T1]{fontenc}    
\usepackage{hyperref}       
\usepackage{url}            
\usepackage{booktabs}       
\usepackage{amsfonts}       
\usepackage{nicefrac}       
\usepackage{microtype}      
\usepackage{lipsum}
\usepackage{fancyhdr}       
\usepackage{graphicx}    
\usepackage{caption}
\usepackage{natbib}
\usepackage{tablefootnote}

\newcommand{\beginsupplement}{%
    \renewcommand{\thefigure}{S\arabic{figure}}%
    \renewcommand{\figurename}{Supplementary Figure}%
    \setcounter{figure}{0}%
}

\usepackage{amsmath}
\usepackage{amssymb}
\usepackage{algorithm}
\usepackage[noend]{algpseudocode}

\usepackage{url}

\usepackage{breakurl}

\graphicspath{{media/}}     

\title{ConSol: Sequential Probability Ratio Testing to Find Consistent LLM Reasoning Paths Efficiently}

\author{
  Jaeyeon Lee, Guantong Qi, Matthew Brady Neeley, Zhandong Liu\thanks{Co-corresponding authors}, Hyun-Hwan Jeong\footnotemark[1] \\
  Baylor College of Medicine \\
  Houston, Texas, USA\\
  \texttt{\{jaeyeon.lee, guantong.qi, matthew.neeley, zhandonl, hyun-hwan.jeong\}@bcm.edu} \\
}

\begin{document}
\maketitle

\begin{abstract}
Recent advancements in large language models (LLMs) integrating explicit reasoning, such as OpenAI's o3-mini, DeepSeek-R1, and QWQ-32B, enable smaller models to solve complex tasks by generating intermediate reasoning steps prior to providing answers. However, this approach significantly increases computational costs, both monetarily and environmentally. The widely-used self-consistency method further exacerbates these costs by aggregating multiple reasoning paths to improve accuracy, often requiring between 40 to 64 samples per task. Although aggregation effectively reduces variance and bias, additional sampling can lead to diminishing returns when early samples yield consistent results. To address inefficiencies, we propose leveraging Sequential Probability Ratio Testing (SPRT) to dynamically terminate sampling once sufficient consistency is achieved. We calibrate SPRT parameters specifically for LLM applications, accounting for sensitivity to detect the mode of the distribution. Our experiments demonstrate that incorporating SPRT significantly enhances token efficiency, achieving comparable accuracy to self-consistency methods but at a substantially reduced computational cost. To promote transparency and facilitate reproducibility, we have made the source code and datasets used in our experiments publicly available at our GitHub repository: \url{https://github.com/LiuzLab/consol}, or available as a PyPI package: \texttt{pip install consol}. We hope that this resource will support further research and encourage the development of new methods building upon our work.

\end{abstract}


\begin{table}[b]
\caption{A Comparison of Accuracy (Top) and Token Reduction (Bottom)}
\centering
\begin{tabular}{lccccc}
\toprule

Method & Self-Cons@40 & p-value \tablefootnote{Sequential Testing with p-value, detailed in Section \ref{sec:methods-baselines}} & Ada-Cons & SPRT (Ours) & mSPRT (Ours) \\
Benchmark / LLM &  &  &  &  &  \\
\midrule
AIME24 / o3-mini-low & 70.0\% & 70.0\% & 70.0\% & 80.0\% & 76.7\% \\
AIME24 / o3-mini-medium & 90.0\% & 93.3\% & 90.0\% & 90.0\% & 90.0\% \\
AIME24 / o3-mini-high & 96.7\% & 96.7\% & 96.7\% & 96.7\% & 96.7\% \\
GPQA Diamond / o3-mini-low & 68.2\% & 69.2\% & 67.2\% & 66.7\% & 70.2\% \\
GSM / gpt-4o-mini & 94.2\% & 94.2\% & 94.2\% & 93.5\% & 93.9\% \\
\midrule
\midrule
AIME24 / o3-mini-low & 0.0\% & 49.3\% & 47.9\% & \bf 63.9\% & 57.3\% \\
AIME24 / o3-mini-medium & 0.0\% & 60.2\% & 62.8\% & \bf 84.8\% & 76.8\% \\
AIME24 / o3-mini-high & 0.0\% & 54.0\% & 64.6\% & \bf 84.5\% & 82.2\% \\
GPQA Diamond / o3-mini-low & 0.0\% & 63.2\% & 70.1\% & \bf 86.2\% & 83.6\% \\
GSM / gpt-4o-mini & 0.0\% & 79.2\% & 82.8\% & 88.6\% & \bf 88.7\% \\
\bottomrule
\end{tabular}\label{tab:table-comparable-accuracy-with-significantly-reduced-cost}
\end{table}

\section{Introduction} \label{sec:introduction}

The recent development of large language models (LLMs) with integrated reasoning processes—such as OpenAI's o1 and o3-mini \citep{openai_introducing_2024, openai_openai_2025}, DeepSeek-R1 \citep{guo_deepseek-r1_2025}, and QWQ-32B \citep{QWQ32B}—has enabled tackling complex problems (e.g., AIME24) using smaller models by generating reasoning tokens prior to producing a final answer. However, this approach incurs substantial computational costs, both in monetary terms and environmental impact (e.g., CO$_{2}$ emissions) \citep{JMLR:v24:23-0069}, but also in terms of significant runtime overhead, which becomes particularly problematic when running large batches of queries with reasoning LLMs on consumer hardware.
For instance, while DeepSeek-R1-1.5B/8B demonstrates performance comparable to that of Llama 70B/450B and Qwen-2.5-32B, it requires generating a larger number of tokens to achieve similar results (\autoref{sub_fig:boxplot_aime}). Because costs are frequently proportional to the number of processed tokens, token efficiency has become a critical factor.

LLMs also incorporate diversity in their reasoning paths, as demonstrated by \citet{wang_self-consistency_2023}. The self-consistency approach assumes that accurate reasoning paths are likely to appear among multiple samples, thus enhancing accuracy through aggregating predictions from a fixed number of samples. Many recent studies have adopted self-consistency in reporting performance on reasoning benchmarks, a strategy that typically increases computational costs by a factor of 40 to 64 times \citep{openai_openai_2025, guo_deepseek-r1_2025}.

From a statistical standpoint, aggregating multiple samples reduces both variance and bias, leading to more stable estimates. The final answer is commonly determined as the mode of the reasoning paths (often based on 40 or 64 samples). In accordance with the law of large numbers, variance decreases as more samples are incorporated, thereby suppressing outliers and reinforcing probable solutions. This iterative process improves statistical robustness and reduces estimation error, ultimately boosting model accuracy.

Nevertheless, the number of samples needed to enhance accuracy depends on the consistency of the LLMs’ responses. When initial samples consistently yield the same answer, additional sampling may offer diminishing returns. In these instances, early termination of the sampling process can optimize token efficiency without sacrificing accuracy.

Adaptive Consistency, as proposed by \citet{aggarwal_lets_2023}, aims to reduce the required number of samples by sequentially testing Bayesian Beta/Dirichlet posteriors, thereby enabling early stopping. However, their stopping rule does not account for Type I error, which risks inflating the false positive rate due to repeated "peeking" at the data during sequential sampling \citep{stefan_big_2023}.

To mitigate the inflation of Type I error in a sequential testing framework—where the sample size is not predetermined but data are analyzed continuously until a significant result is detected—several statistical methods have been proposed, including the Sequential Probability Ratio Test (SPRT) and its variants \citep{wald_sequential_1945}. These techniques are widely used in clinical trials to confirm the significant impact of medical treatments and to reduce financial costs by terminating the test early. Despite their proven effectiveness in clinical settings, to the best of our knowledge, such methods have not yet been applied to improve LLM accuracy and efficiency in reasoning benchmarks.

In this paper, our contribution includes:

\begin{itemize}

\item {\bf Formulating a Statistical Framework}: We introduce a novel categorical distribution modeling to leverage Sequential Probability Ratio Testing (SPRT) techniques to efficiently identify consistent answers.
\item {\bf Calibrating SPRT Parameters for LLMs}: We calibrated the SPRT parameters to find a mode of the categorical distribution of LLM responses by accounting for sensitivity to small probabilistic differences and tolerating a certain rate of Type II errors.
\item {\bf Enhancing Token Efficiency}: We demonstrate that our approach achieves improved token efficiency while maintaining accuracy on par with Self-Consistency methods. (See \autoref{tab:table-comparable-accuracy-with-significantly-reduced-cost})
\end{itemize}

\section{Proposed Method}
\subsection{Problem Formulation}

According to the hypothesis in Self-Consistency, when a reasoning model is repeatedly queried to solve a reasoning task, each individual reasoning path is generated with some diversity. However, correct reasoning processes tend to converge on the same final answer more often than incorrect ones. 

To leverage a statistical approach based on this hypothesis, we propose to assume that the final output of the LLM is sampled independently from a categorical distribution over the set of possible answers. Specifically, each sample is drawn from the distribution:
\[
Y_i \sim \text{Categorical}(\boldsymbol{p}), \quad \text{for } i = 1, \dots, N
\]
Here, \(Y_i \in \{Answer_1,Answer_2,\dots,Answer_k\}\) is the observed answer for the \(i\)th sample, and \(\boldsymbol{p} = (p_1, p_2, \dots, p_k)\) is the fixed probability vector over \(k\) possible answer categories. 

For each answer category \(Answer_i\), count how many times it appears:
\[
n_i = \sum_{j=1}^N \mathbf{1}\{Y_j = Answer_i\}.
\]

The maximum likelihood estimate (MLE) for the probability \(p_i\) is given by:
\[
\hat{p}_i = \frac{n_i}{N}.
\]

This is because the relative frequency of each answer converges to the true probability \(p_i\) as \(N\) increases by the law of large numbers.

The mode is the answer category with the highest estimated probability:
\[
\hat{i}_{\text{mode}} = \operatorname{argmax}_{i} \hat{p}_i.
\]
Equivalently, this is the category with the maximum count \(n_i\).

As the number of samples \(N\) grows, the empirical frequencies \(\hat{p}_i\) converge to the true probabilities \(p_i\), making \(\hat{i}_{\text{mode}}\) a consistent estimator of the true mode \(i^* = \operatorname{argmax}_{i} p_i\). 

In this proposed scheme of assuming categorical distribution for LLM outputs, the statistical power of the selection of the number of samples for Self-Consistency can be interpreted with Hoeffding's inequality\footnote{By applying Hoeffding’s inequality to the empirical frequency of each of the 4 categories and using a union bound, we can derive that to have at least 95\% confidence that the deviation of the observed frequency from the true probability is no more than 0.25, the number of samples \( n \) must satisfy $4 \cdot 2\exp(-2n(0.25)^2) \leq 0.05.$. Solving this yields a conservative estimate of approximately 40 samples. Note that this is a simplified analysis that assumes independent samples and a fixed difference between the top probabilities.}. And also it enables us to employ solid and established approaches to minimize the number of sample sizes while satisfying Type I and II error constraints, as described in Appendix \ref{sec:appendix-sprt}.

\subsection{Simplification to Bernoulli Distribution of Two Most Frequent Responses} \label{sec:methods-simplification}

\begin{figure}[h]
\includegraphics[width=1\textwidth]{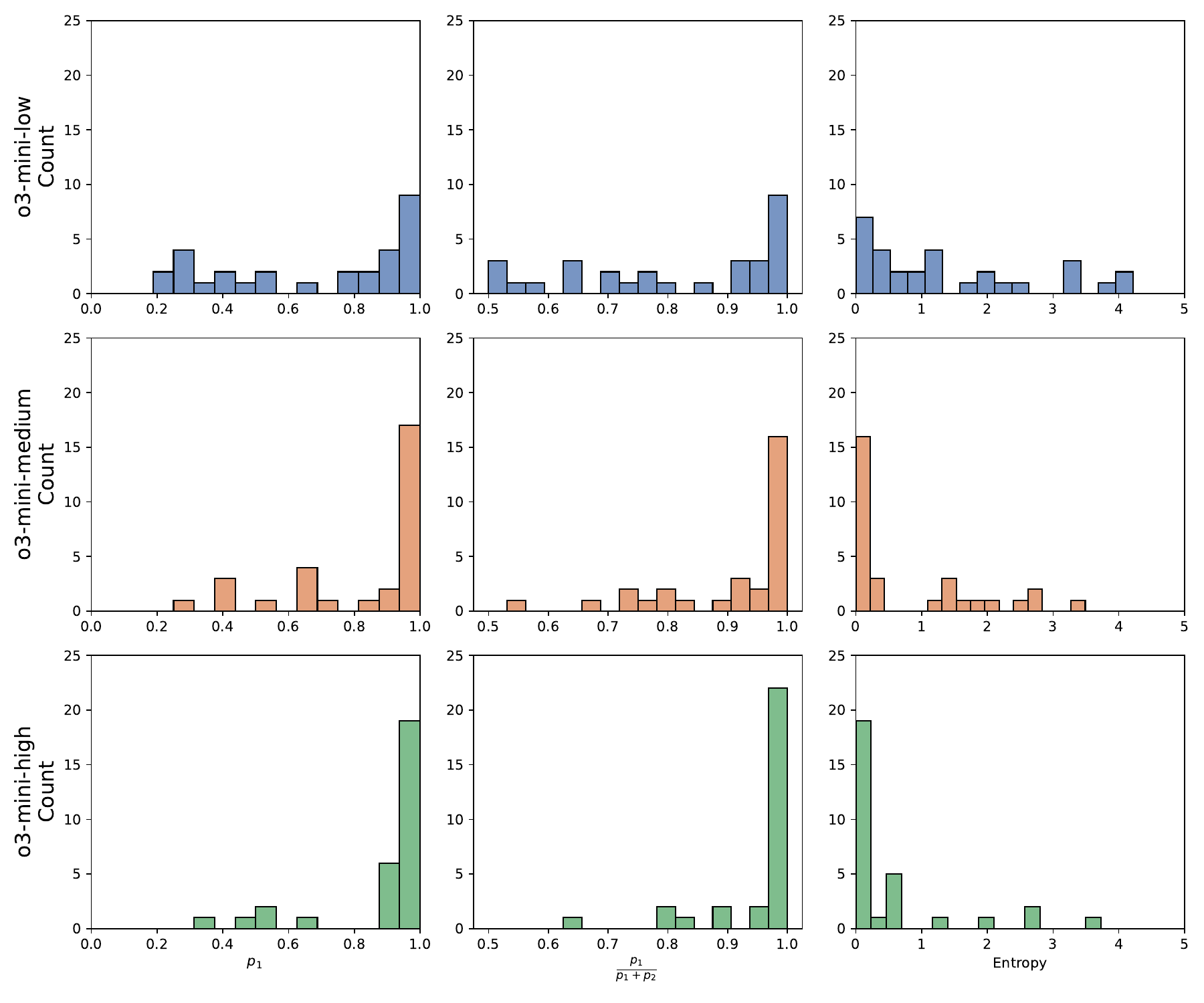}
\caption{Distribution of Probabilities and Entropy for OpenAI's o3-mini Reasoning Models' Responses.
We observe that a majority of the $p_1$ values are above $0.5$, indicating that a single response often dominates the others. This tendency becomes more pronounced when stronger models are used relative to the difficulty of the benchmark (Left). The ratio between $p_1$ and $p_2$ is also skewed to the right. As the models become stronger, dominant responses become even more prevalent (Middle). Finally, entropy decreases with model strength, indicating reduced randomness in responses by stronger models (Right).
Note: the distributions are based on 40 samples.}
\label{fig:figure-histogram-for-data-distribution}
\end{figure}

Empirical observations of LLM reasoning tasks reveal that one response typically emerges as dominant compared to the other alternatives (see \autoref{fig:figure-histogram-for-data-distribution}). This observation motivates a simplification of our analysis: rather than modeling the full categorical distribution over all $k$ responses, we focus on the two most frequent outcomes. This simplification was also practiced by \citet{aggarwal_lets_2023}, where they simplified the Dirichlet distribution to the Beta distribution. This reduction allows us to approximate the original model with a Bernoulli distribution while preserving the sufficient information needed for hypothesis testing.

Let \( p_{\text{first}} \) and \( p_{\text{second}} \) denote the true probabilities of the most frequent and second-most frequent responses, respectively. We define the simplified success probability \( p' \) as:
\[
p' = \frac{p_{\text{first}}}{p_{\text{first}} + p_{\text{second}}}.
\]
Under the assumption of no dominant response, these two probabilities would be equal, yielding \( p' = 0.5 \). This naturally leads to the following hypothesis test:
\[
H_0: p' = 0.5 \quad \text{versus} \quad H_1: p' > 0.5.
\]
Here, the null hypothesis \( H_0 \) asserts that the two most frequent responses occur with equal likelihood, whereas the alternative hypothesis \( H_1 \) captures the empirically observed imbalance, indicating that one response is significantly more likely.

By recasting the mode estimation as a one-tailed Bernoulli experiment, we can employ standard statistical methods to assess whether the observed deviation from the 0.5 baseline is statistically significant. A significant result, that is, when \( p' \) is found to be considerably greater than 0.5, provides strong evidence of a dominant answer. This not only validates our Bernoulli approximation, but also enables the use of established techniques to determine the minimal number of samples required while controlling for Type I and II errors, as further discussed in Section \ref{sec:introduction}.

\subsection{Sequential Testing Methods of Two Most Frequent Responses}  \label{sec:methods-sequential-testing}

By modeling the two most frequent responses as a Bernoulli process, we can leverage sequential testing to efficiently determine if one answer truly dominates. This approach allows the test to continuously monitor incoming data and terminate early when sufficient evidence is gathered, thereby reducing unnecessary computation, while still rigorously controlling Type I and Type II errors. As the sequential test compares the observed success probability against the baseline of $0.5$, it dynamically adapts to the strength of the emerging signal. This not only enhances computational efficiency but also provides a statistically robust method for identifying a dominant response in LLM outputs, ensuring that the decision-making process is both timely and reliable.

\subsubsection{Sequential Probabilistic Ratio Testing (SPRT)}

Consider the two most frequent responses—“first” and “second”—with a total of
\[
n' = n_{\text{first}} + n_{\text{second}}
\]
samples, where \(n_{\text{first}}\) is the number of times the first response is observed (the “dominant” response) and \(n_{\text{second}}\) is the number of times the second response is observed.

We define the simplified success probability as:
\[
p' = \frac{p_{\text{first}}}{p_{\text{first}}+p_{\text{second}}},
\]
and set up the hypothesis:
\[
H_0: p' = 0.5 \quad \text{versus} \quad H_1: p' = 0.5001.
\]
Under \(H_0\), both responses are equally likely, while under \(H_1\) the first response is slightly more likely.

For each observation, define an indicator variable \(X\) such that:
\[
X = \begin{cases}
1, & \text{if the response is the first (dominant) response}, \\
0, & \text{if the response is the second}.
\end{cases}
\]
The likelihood for an observation under a given hypothesis is then:
\[
L(p'; X) = (p')^X (1-p')^{1-X}.
\]
Thus, the likelihood ratio for an observation is:
\[
\lambda = \frac{L(0.6; X)}{L(0.5; X)} =
\begin{cases}
\frac{0.6}{0.5} = 1.2, & \text{if } X=1, \\
\frac{0.4}{0.5} = 0.8, & \text{if } X=0.
\end{cases}
\]

After collecting \(n'\) samples, the cumulative likelihood ratio is:
\[
\Lambda_{n'} = (1.2)^{n_{\text{first}}} \, (0.8)^{n_{\text{second}}}.
\]

Based on the calculated \(\Lambda_{n'}\), the following steps are taken:

\begin{enumerate}
  \item If \(\Lambda_{n'} \geq A\): Reject \(H_0\) (i.e., conclude that the first response is significantly dominant).
  \item If \(\Lambda_{n'} \leq B\): Accept \(H_0\) (i.e., conclude that there is no significant dominance).
  \item Otherwise: Continue sampling additional responses.
\end{enumerate}

Here, \(A\) and \(B\) are the predetermined thresholds, as explained in Appendix \ref{sec:appendix-sprt}.

\subsubsection{Mixture SPRT}

As explained in Section 2, the mixture SPRT adapts the classical SPRT by integrating over a range of plausible values for the alternative hypothesis rather than fixing it at a single value. In our setting, we wish to test
\[
H_0: p' = 0.5 \quad \text{versus} \quad H_1: p' > 0.5,
\]
where \(p'\) is the probability of observing the first response. Here, \(n'\) denotes the total number of samples (i.e., the combined number of first and second responses), \(S_{\text{first}}\) is the number of samples corresponding to the first response, and \(S_{\text{second}}\) is the number of samples corresponding to the second response. 

A natural choice is to place a truncated Beta prior over \(p'\) on the interval \((0.5, 1]\). We denote this prior as
\[
\pi(p') \propto \text{Beta}(p'; \alpha_0=10^6, \beta_0=10^6) \quad \text{for } p' \in (0.5, 1],
\]
where \(\alpha_0\) and \(\beta_0\) are hyperparameters that encode our prior belief about the effect size under the alternative. The truncation to \((0.5, 1]\) reflects the belief that any deviation from the null must be in the direction of a higher probability for the first response.

The mixture likelihood ratio then becomes
\[ \Lambda_{n'} = \frac{\int_{0.5}^{1} L(p') \, \pi(p') \, dp'}{L(0.5)} = \frac{\int_{0.5}^{1} p'^{\, n_{\text{first}}}(1-p')^{\, n_{\text{second}}} \pi(p') \, dp'}{(0.5)^{\, n_{\text{first}}}(0.5)^{\, n_{\text{second}}}} \]
which averages the likelihood over the range of alternative values, weighted by the prior. As in the classical SPRT, the decision is made by comparing the resulting \(\Lambda_{n'}\) with the thresholds \(A\) and \(B\) chosen to control the overall error rates.

\subsection{Parameter Adjustment for Practical LLM Application} \label{sec:methods-parameter-adjustment}

While conventional parameter settings for SPRT and mSPRT exist to ensure statistical guarantees on both Type I and Type II errors in typical clinical applications, these settings often require sample sizes that exceed practical limits in our specific application context. Therefore, we adapted the parameters to balance statistical rigor and practical feasibility, prioritizing sufficient accuracy for our particular use case.

\subsubsection{Enhancing Sensitivity to Small Probability Differences}

Sequential testing is commonly used to evaluate significant probability increments resulting from treatments, with standard parameter configurations well-established for these scenarios. However, our original objective involves identifying the most frequent category within a categorical distribution. In such cases, even minimal probability differences between the two leading responses can be critical.

In SPRT applications for Bernoulli processes, sensitivity is strongly influenced by the proximity of the parameters \(p_0\) and \(p_1\). Closer values of these parameters demand larger sample sizes to reliably detect small differences. Standard parameter values typically range between \(0.55\) and \(0.6\), but we reduced \(p_1\) to \(0.5001\) to significantly heighten sensitivity to marginal differences.

For the mixture SPRT, the sensitivity is governed by the Beta prior distribution. Standard practice employs non-informative priors, such as a Uniform prior (\(\text{Beta}(p'; \alpha_0=1, \beta_0=1)\)) or Jeffreys prior (\(\text{Beta}(p'; \alpha_0=0.5, \beta_0=0.5)\)). These conventional priors assume either a flat probability density across \([0, 1]\) or a bimodal, U-shaped distribution emphasizing extremes at \(0\) and \(1\). To enhance sensitivity specifically around a probability near \(0.5\), we employed a symmetric, bell-shaped, unimodal prior distribution: \(\text{Beta}(p'; \alpha_0=10^6, \beta_0=10^6)\).

\subsubsection{Compromising Type II Error with Extremely High $\beta$}

\begin{figure}[h]
\includegraphics[width=1\textwidth]{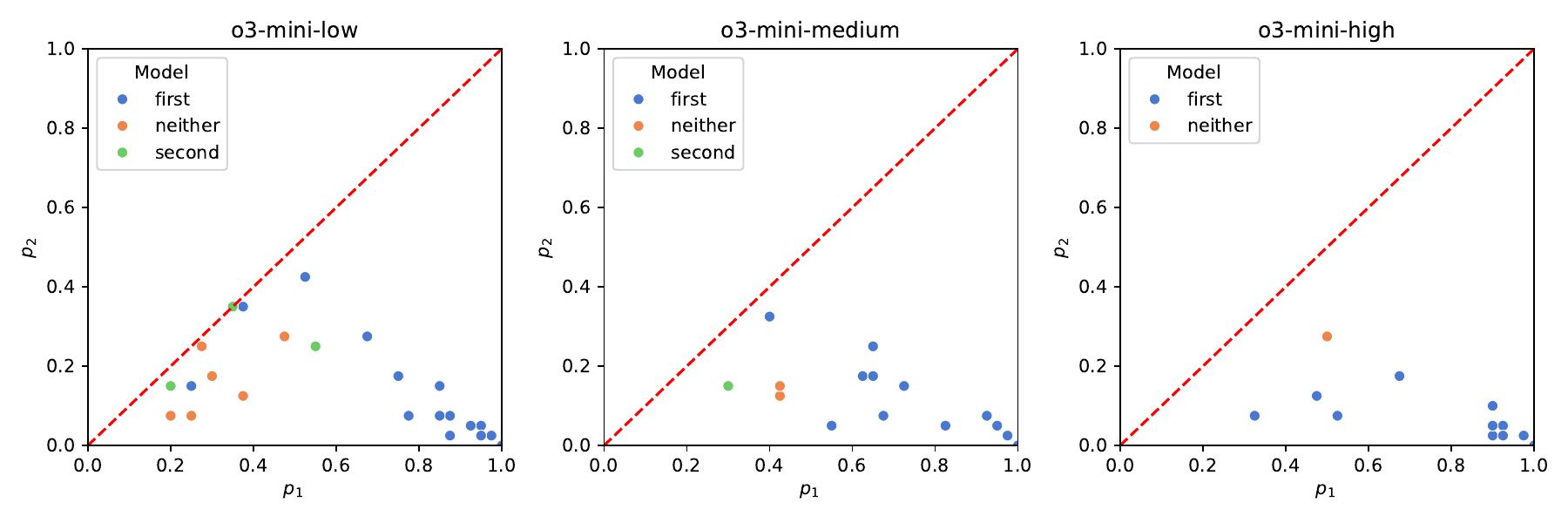}
\caption{Scatterplot for the two most frequent responses by Open AI o3-mini reasoning models. The color represents the correctness of the most frequent answers. When \( p_1 \) is less than 0.5, it's unlikely that either of the two most frequent responses is the correct answer. This observation justifies our approaches to allow early stopping by accepting $H_0$, concluding there's no dominating response, even with a weak evidence.}
\label{fig:figure-scatterplot-for-top-two-probabilities}
\end{figure}

Both $\alpha$ and $\beta$ are shared parameters for both SPRT and mixture SPRT, and conventionally have very low values, for example, $0.05$, and $0.10$ to manage Type I and Type II errors, respectively. In our problem setting, high tolerance to Type I error, $\alpha$, means early stopping without enough evidence to believe the most frequent response is dominant. Meanwhile, high tolerance to Type II error, $\beta$, means early stopping without enough evidence to believe there is no dominating response. Both are obviously errors that we should avoid, but the risk is not equal for the original purpose of finding the mode.

Because, as shown in \autoref{fig:figure-scatterplot-for-top-two-probabilities} and \autoref{sub_fig:scatter}, most of the modes are highly dominant. When a mode is not dominant, its corresponding response typically exhibits lower accuracy, making further sampling less valuable. To address this, we allowed the sampling process to end early - even with weak evidence - when observations indicate that there is no dominant response, setting an extremely high value for the parameter $\beta \approx 1 - \alpha$. This approach drastically reduces the number of samples while maintaining a low Type I error rate during the process.

\subsection{Algorithms and Implementation Details}

Our algorithm follows the sample-by-sample approach introduced by \citet{aggarwal_lets_2023}, with notable modifications. Instead of sequentially sampling one response at a time, we identify the minimum number of samples required before breaking the loop and then concurrently query multiple samples from the LLM. This straightforward enhancement leverages concurrency, allowing simultaneous sampling of multiple LLM outputs, significantly reducing the running time.

This concurrent querying approach is particularly advantageous when using cloud-based LLM platforms—such as OpenAI, Gemini, and DeepSeek—which efficiently handle thousands of concurrent requests. Typically, our method concurrently samples two or three responses per turn, as illustrated in \autoref{sub_fig:figure-boxplot-for-concurrency-for-each-turn.pdf}, resulting in substantial reductions in overall runtime.

\begin{algorithm}
\caption{Sequential Testing}
\begin{algorithmic}[1]
\Function{SequentialTesting}{P, N} \Comment{$P$: the problem, $N$: the number of max trials}
    \State $O \gets \{\}$ \Comment{O: List to store all outputs}
    \While{True}
        \State $(first,\, second) \gets$ \Call{GetTopTwoFrequentOutputs}{$O$}
        \State $K \gets$ \Call{DetermineTrial}{$first,\, second,\, N,\, O$} \Comment{$K$: Minimal trials required in the best-case scenario}
        \If{$K = 0$}
            \State break
        \EndIf
        \For{$i \gets 1$ \textbf{to} $K$} \Comment{Parallelized for concurrency}
            \State  $O \gets O \cup \{\Call{RunLLMSolver}{P}\}$
        \EndFor
    \EndWhile
    \State \Return $first$
\EndFunction
\end{algorithmic}
\end{algorithm}

\begin{algorithm}
\caption{A Function to determine the minimal trials required in the best-case scenario}
\begin{algorithmic}[1]
\Function{DetermineTrial}{first, second, N, O} 
    \For{$T \gets 0$ \textbf{to} $N$}
        \If{$T + first > 0$}
            \If{ \Call{Test}{$first + T$, $second$} }
                \State \textbf{break}
            \EndIf
        \EndIf
    \EndFor
    \State $T \gets \min(T, N - |O|)$
    \State \Return $T$
\EndFunction
\end{algorithmic}
\end{algorithm}

We used LangChain (\cite{chase_langchain_2022}) as an interface between our algorithm and the LLM model API. The structured output feature was used to standardize the answer format. For probabilistic and statistical calculations, we utilized the SciPy package (\cite{virtanen_scipy_2020}), which provides both an intuitive interface and efficient computation.

\subsection{Baselines} \label{sec:methods-baselines}

\subsubsection{Self-Consistency (Self-Cons@40)}

Self-Consistency enhances the accuracy of Large Language Models (LLMs) by selecting the most consistent answer from multiple reasoning path samples of a predetermined size. Typically, sample sizes of 64 \citep{wang_self-consistency_2023, openai_openai_2025, guo_deepseek-r1_2025} or 40 \citep{aggarwal_lets_2023} have been used in prior studies. For simplicity and comparability, we adopt a sample size of 40, aligning our methodology with Adaptive Consistency.

\subsubsection{Adaptive Consistency (Ada-Cons)}

Adaptive Consistency shares similarities with our approach, primarily aiming to reduce computational cost through early stopping based on statistical evaluation of answer consistency. Adaptive Consistency addresses this by applying Bayesian inference methods, including the Chinese Restaurant Process, Dirichlet distribution, and Beta distribution, to assess confidence intervals at a $95\%$ confidence level to conclude earlier than 40 maximum number of samples. Since all three variants showed comparable performance, we selected the Beta distribution variant for our comparative analysis. A critical challenge in sequential hypothesis testing is controlling the cumulative Type I error rate, as repeated hypothesis tests without proper adjustments can inflate false-positive rates, which will be further detailed in the next section.

\subsubsection{Sequential Testing with p-value}

Additionally, we propose a sequential hypothesis testing baseline employing frequentist p-values, analogous to the Adaptive Consistency method. Following the simplification described in Section \ref{sec:methods-simplification}, we model consistency evaluation as a Bernoulli process and thus employ the Binomial distribution:

We define:
\begin{align*}
    N &= n_{\text{first}} + n_{\text{second}},\\
    K &= n_{\text{first}},\\
    K &\sim \text{Binomial}(N, p')
\end{align*}

The hypotheses are formally defined as:
\[
H_0: p' = 0.5 \quad \text{versus} \quad H_1: p' > 0.5.
\]

The p-value is computed as the probability of observing at least \(K\) successes under the null hypothesis:
\[
    \text{p-value} = P_{H_0}(X \ge K) = \sum_{k=K}^{N} \binom{N}{k} \left(\frac{1}{2}\right)^N = 1 - \text{BinomialCDF}(K-1; N, 0.5).
\]

We set a significance level (\(\alpha\)) of 0.05 and reject \(H_0\) when \(\text{p-value} < \alpha\). This method serves as a frequentist counterpart to Adaptive Consistency.

Despite the intuitiveness, a notable consideration with sequential hypothesis testing using p-values is the potential inflation of the cumulative Type I error rate due to repeated interim analyses. Each test conducted during the sequential process introduces additional opportunities for false positives.  This practice, called `data peeking' or `optional stopping', happens when researchers repeatedly check their data and stop collecting more once they find a significant result 
 (\cite{stefan_efficiency_2022}). Doing this increases the risk of false positives, especially if they check too often or add only a few new data points at a time. The more frequently they analyze, the higher the chance of mistakenly finding a significant effect.

To mitigate this issue, it is essential to implement corrections, such as the Bonferroni adjustment or advanced sequential testing frameworks like Wald’s Sequential Probability Ratio Test (SPRT), as our proposed method. Without these adjustments, naive sequential application of p-value tests may yield overly permissive decisions, thereby jeopardizing the validity and reliability of the resulting conclusions.

\section{Experiments}

\subsection{Setting up the Benchmarks}

Our approach is LLM-model-agnostic, meaning it is applicable to any LLM model or benchmark where self-consistency methods enhance performance. This allows us to achieve the same accuracy at a lower cost. We primarily focus on reasoning LLM models and benchmarks, conducting our experiments first on AIME24 and GPQA Diamond \citep{rein2023gpqagraduatelevelgoogleproofqa}. We selected these benchmarks because LLM reasoning models have shown significant improvement on these datasets \citep{openai_introducing_2025}.

We initially use OpenAI's o3-mini with the low reasoning effort option. Following that, we apply medium and high reasoning efforts specifically for the AIME24 dataset, as research indicates that varying levels of reasoning effort can significantly enhance accuracy \citep{openai_openai_2025}. 

Additionally, we conduct another benchmark using GSM-8K \citep{cobbe2021trainingverifierssolvemath} with the GPT-4o-mini model. This choice of LLM model is informed by prior studies on datasets released before the introduction of reasoning models \citep{wang_self-consistency_2023, aggarwal_lets_2023}. Those studies demonstrated that accuracy improved consistently regardless of the reasoning effort employed, and there have been no reports suggesting that reasoning models could enhance accuracy for this particular benchmark.

\subsection{Metrics}

\subsubsection{Efficiency}

Our method enhances accuracy by identifying consistency while minimizing token usage. We quantify token efficiency through Token Reduction, defined as:

\[
\text{Token Reduction} = \frac{T_{\text{Self-Cons@40}} - T}{T_{\text{Self-Cons@40}}} \times 100,
\]

where \(T\) denotes the total tokens utilized by the comparison method, and \(T_{\text{Self-Cons@40}}\) indicates the total tokens used by self-consistency with 40 samples. Higher Token Reduction values signify that the comparison method achieves similar results with fewer tokens.

\subsubsection{Consistency} \label{sec:experiments-consistency}

Since consistency does not always guarantee accuracy improvement, we conducted a simulation study with synthetic data to validate our method's effectiveness. Specifically, we estimated parameters for a categorical distribution based on 40 samples and evaluated our methods alongside baseline approaches in identifying the distribution mode. LLM queries were simulated via sampling from this distribution, allowing us to measure accuracy relative to the estimated distribution mode. We term this metric the Consistency Score to differentiate it clearly from ground-truth accuracy.

\subsection{Parameter Settings}

\subsubsection{For Baseline Methods}

The performance of Self-Consistency, Adaptive-Consistency, and p-value methods depends significantly on threshold selections (confidence/p-value) and the maximum number of generations. For comparability, we standardized the maximum number of generations at 40. The confidence threshold for Adaptive-Consistency was set at 95\%, and the significance level for the p-value method was fixed at 0.05. Detailed analyses of how performance varies with threshold selection are provided in Section \ref{sec:results-consistency}.

\subsubsection{For Proposed Methods}

As outlined in Sections \ref{sec:methods-sequential-testing} and \ref{sec:methods-parameter-adjustment}, parameters for SPRT and mSPRT were calibrated to enhance sensitivity to small probabilistic differences and to allow for controlled Type II error. Specifically, we set parameters to \(p_1 = 5.001\) with \(\beta = 0.949976\) for SPRT, and \(\alpha_0 = \beta_0 = 10^6\) with \(\beta = 0.94994\) for mSPRT. Comprehensive assessments regarding sensitivity to the \(\beta\) parameter are detailed in Section \ref{sec:results-consistency}.

Given that a high $\beta$ value inherently regulates the number of trials, we increased the maximum number of generations to 256, enabling more extensive sampling for potentially inconsistent answers. Nevertheless, in all experiments, early stopping consistently occurred well before reaching this limit.

\section{Results}

\subsection{Comparable Accuracy with Significantly Reduced Costs}


In general, accuracy values are comparable across most methods, with SPRT and mSPRT methods demonstrating competitive or improved performance on certain benchmarks, such as the AIME24 at the \texttt{o3-mini-low} setting, where SPRT shows a notable improvement from 70.0\% to 80.0\%, beyond the accuracy of Self-Consistency of 40 samples. We believe this outcome resulted from allowing a collection of more than 40 samples, which enabled concentration of its efforts on finding consistent answers for questions with high uncertainty (see Appendix \ref{sec:appendix-example-msprt}).


We observe that our method achieves efficiency gains by significantly reducing the number of tokens (see \autoref{tab:table-comparable-accuracy-with-significantly-reduced-cost}).
Notably, the SPRT and mSPRT methods consistently outperform baseline methods by achieving significantly higher token reduction percentages, despite being allowed to sample more than 40. Specifically, SPRT attains the highest token reduction in most cases, such as 84.8\% for AIME24 \texttt{o3-mini-medium} and 86.2\% for GPQA Diamond \texttt{o3-mini-low}, emphasizing its efficiency advantage. The GSM benchmark with \texttt{gpt-4o-mini} sees a slight edge for mSPRT, achieving the highest reduction at 88.7\%.

\subsection{Early Confirmation of Consistency than Adaptive Consistency} \label{sec:results-consistency}

\begin{figure}[h]
\includegraphics[width=\textwidth]{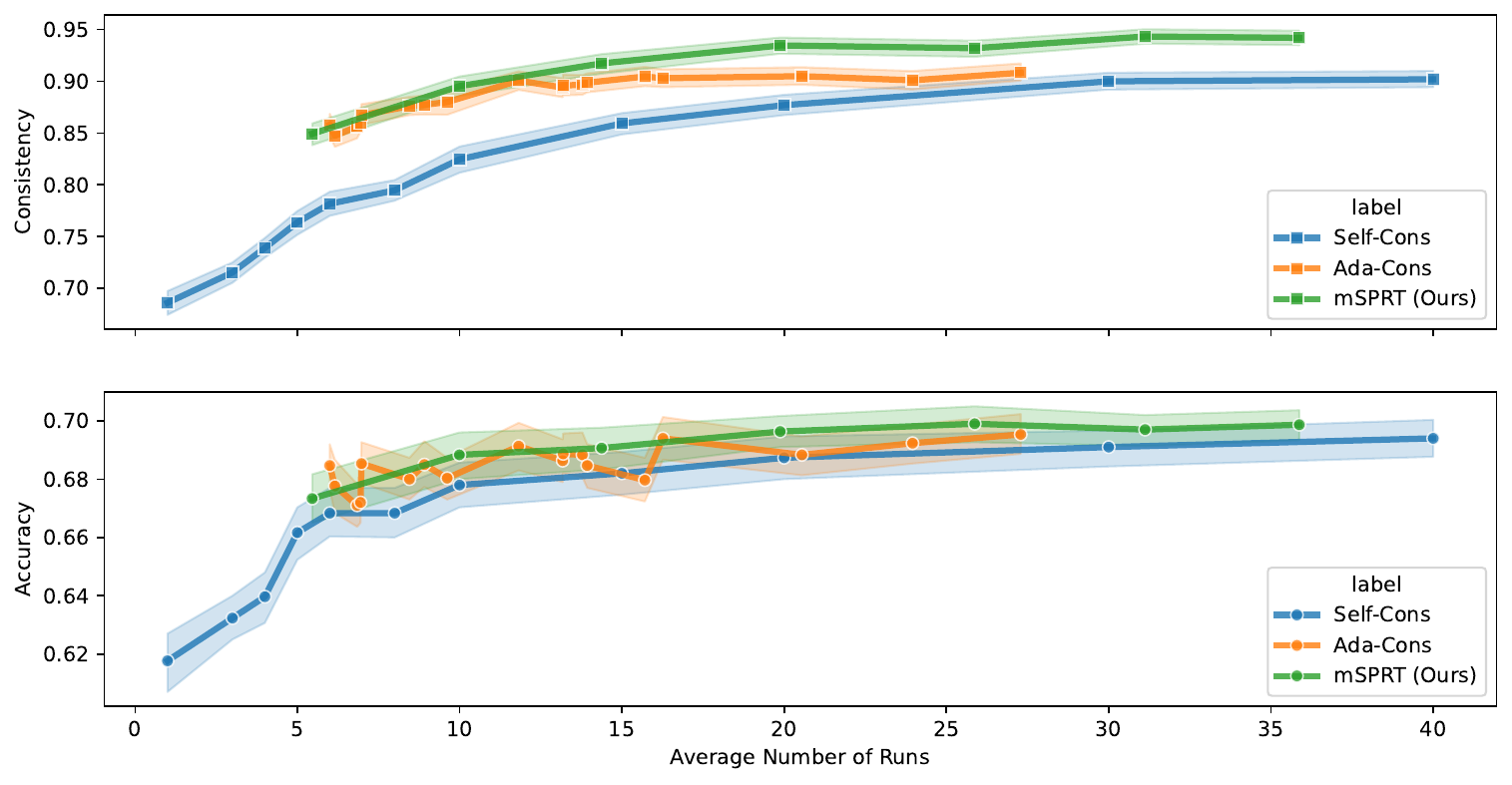}
\caption{Comparison of consistency and accuracy for different methods on the AIME24 benchmark with the o3-mini-low model as a function of the average number of runs. For Self-Consistency, the number of samples varies between \(1\) and \(40\). For Adaptive-Consistency, the confidence threshold parameter ranges from \(0.74\) to \(0.9999\). For Mixture of SPRT, the parameter \(\beta\) ranges from \(0.94979\) to \(0.94997\).}
\label{fig:figure-consistency}
\end{figure}

In the simulation study conducted with synthetic data, as detailed in Section \ref{sec:experiments-consistency}, the proposed method consistently achieves the same level of performance with fewer samples on average compared to Self-Consistency and Adaptive-Consistency (see \autoref{fig:figure-consistency}). Notably, the consistency score of the proposed method is not restricted by the maximum consistency score achieved by Self-Consistency. This is due to the proposed method's capacity to extend beyond the fixed sample size limit of 40, enabling it to more effectively identify the true mode of the underlying synthetic data distribution.








\section{Discussion and Conclusion}

Ensembling techniques such as voting and output aggregation have long been employed to boost the performance of classifiers in machine learning. While self-consistency methods have successfully leveraged majority voting to improve the accuracy of large language models on reasoning tasks, they significantly increase computational costs by requiring multiple reasoning samples. Our study addresses this challenge by formulating a statistically rigorous framework using SPRT tailored explicitly for LLM reasoning outputs. Unlike existing approaches, including Adaptive Consistency, our method carefully controls Type I error to prevent false positives resulting from continuous analysis of sequential data.

By modeling the reasoning outputs as categorical distributions, we calibrated SPRT parameters specifically for finding a mode of the distribution of LLM responses, balancing sensitivity to probabilistic differences and tolerance for Type II errors. This approach allows early termination of sampling once sufficient statistical evidence for a dominant answer is established, significantly improving token efficiency. As demonstrated by our experimental results, our SPRT-based methods achieves comparable accuracy to traditional self-consistency approaches while substantially reducing computational costs, addressing the critical issue of token efficiency emphasized earlier.

Despite promising results, our method currently faces limitations, including dependence on sensitive parameter settings and the scope of benchmarks evaluated, primarily limited to simpler tasks such as short-answer and multiple-choice questions. Extending our approach to more complex reasoning tasks, including generating detailed consensus-driven responses, remains a priority for future research. Additionally, exploring generalized and systematic calibration methods to reduce sensitivity to parameter settings would further enhance applicability.

To promote transparency and facilitate reproducibility, we have publicly released our source code and datasets via GitHub at \url{https://github.com/LiuzLab/consol}, and as a PyPI package accessible through \texttt{pip install consol}. We believe this resource will support future research and encourage further innovations building upon our proposed statistical framework.

\section*{Acknowledgments}
This research was supported in part by the Cancer Prevention and Research Institute of Texas (CPRIT, grant number RP240131), the Chan Zuckerberg Initiative (grant number 2023-332162), the Chao Endowment and the Huffington Foundation, and Jan and Dan Duncan Neurological Research Institute at Texas Children’s Hospital.


\bibliography{consol}  

\beginsupplement

\newpage
\section*{Appendix}

\renewcommand{\thesubsection}{\Alph{subsection}}

\subsection{Sequential Probability Ratio Test} \label{sec:appendix-sprt}

The Sequential Probability Ratio Test (SPRT), introduced by \cite{wald_sequential_1945}, is renowned not only for its sequential nature but also for its optimality. Under the assumption that the observations are independent and identically distributed (i.i.d.), the SPRT minimizes the average sample number (ASN) required to reach a decision among all tests that satisfy given Type I and Type II error constraints. In other words, for a fixed pair of error probabilities, no other sequential test can achieve a lower average sample size when the true parameter value is either under the null hypothesis $H_0$ or the alternative hypothesis $H_1$.

The test proceeds by computing the likelihood ratio:
\[
\Lambda_n = \prod_{i=1}^n \frac{f(X_i; \theta_1)}{f(X_i; \theta_0)},
\]

and stops when this ratio exceeds an upper threshold $A$ or falls below a lower threshold $B$. The thresholds \(A\) and \(B\) are chosen based on the specified Type I error probability \(\alpha\) and Type II error probability \(\beta\). Typically, they are set as:
\[
A = \frac{1-\beta}{\alpha} \quad \text{and} \quad B = \frac{\beta}{1-\alpha}.
\]

For example, if we set the Type I error probability \(\alpha = 0.05\) and the Type II error probability \(\beta = 0.10\), the thresholds become:
\[
A = \frac{1 - \beta}{\alpha} = \frac{0.90}{0.05} = 18 \quad \text{and} \quad B = \frac{\beta}{1 - \alpha} = \frac{0.10}{0.95} \approx 0.1053.
\]
This implies that the test will stop and reject \(H_0\) if the likelihood ratio \(\Lambda_n\) exceeds 18, and it will accept \(H_0\) if \(\Lambda_n\) falls below approximately 0.1053. When \(\Lambda_n\) remains between these two thresholds, the test continues to collect more observations until a decision can be made.

\subsection{Mixture of Sequential Probability Ratio Test}

The mixture of Sequential Probability Ratio Test (mixture SPRT), or Sequential Bayes Factor Test (SBFT), extends the classical SPRT by incorporating uncertainty about the alternative hypothesis through a mixture model. Instead of a fixed alternative, mSPRT integrates the likelihood over a prior distribution, making it more robust when the true parameter is unknown.

Mathematically, mSPRT evaluates the mixture likelihood ratio:

\[
\Lambda_n = \frac{\int_{\Theta} L(\theta; X_1, \ldots, X_n) \pi(\theta) \, d\theta}{L(\theta_0; X_1, \ldots, X_n)},
\]

where \( L(\theta; X_1, \ldots, X_n) \) is the likelihood, \( \pi(\theta) \) is the prior over \(\Theta\), and \(\theta_0\) is the null hypothesis parameter. Data collection continues until \(\Lambda_n\) crosses predefined thresholds, balancing Type I and II error rates.

\newpage
\subsection{Example Cases of Inconsistent Responses} \label{sec:appendix-example-msprt}

The following example illustrates an inconsistent response obtained while solving problem AIME 2024-II-8 using the LLM model \texttt{o3-mini-low} with the proposed Mixture of SPRT method. The process required 63 samples to reach a conclusion. Given that the frequency difference between the two most common responses is merely 3, insufficient sampling might lead to an incorrect answer.

\subsubsection{Example}

\textbf{Problem Statement:} Torus $T$ is the surface produced by revolving a circle with radius $3$ around an axis in the plane of the circle that is a distance $6$ from the center of the circle (so like a donut). Let $S$ be a sphere with a radius $11$. When $T$ rests on the inside of $S$, it is internally tangent to $S$ along a circle with radius $r_i$, and when $T$ rests on the outside of $S$, it is externally tangent to $S$ along a circle with radius $r_o$. The difference $r_i-r_o$ can be written as $ frac{m}{n}$, where $m$ and $n$ are relatively prime positive integers. Find $m+n$.

\textbf{Answer:} \boxed{127}

\textbf{Solution:} First, let's consider a section $\mathcal{P}$ of the solids, along the axis. By some 3D-Geometry thinking, we can simply know that the axis crosses the sphere center. So, that is saying, the $\mathcal{P}$ we took crosses one of the equator of the sphere.

Here I drew two graphs, the first one is the case when $T$ is internally tangent to $S$, and the second one is when $T$ is externally tangent to $S$.

In both cases, we know $\Delta OEF \sim \Delta OGH \Longrightarrow \frac{EF}{OE} =\frac{GH}{OG}$.

Hence, in the case of internal tangent, $\frac{6}{11-3} =\frac{r_i}{11} \Longrightarrow r_i=\frac{33}{4}$.

In the case of external tangent, $\frac{6}{11+3} =\frac{r_o}{11} \Longrightarrow r_o=\frac{33}{7}$.

Thereby, $r_i-r_o=\frac{33}{4}-\frac{33}{7}=\frac{99}{28}$. And there goes the answer, $99+28=\boxed{127}$.

\hfill
\begin{table}[htbp]
\centering

\begin{tabular}{ll}
\toprule
 & Responses \\
count &  \\
\midrule
9 & \boxed{127} \\
6 & 19, 55, 13 \\
5 & 23, 17 \\
4 & 31 \\
3 & 29, 61 \\
2 & 7, 24 \\
1 & 18, 113, 169, 73, 155, 11, 14, 22, 53, 15 \\
\bottomrule
\end{tabular}

\end{table}



\newpage
\renewcommand{\figurename}{Figure}
\renewcommand{\thefigure}{S\arabic{figure}}

\setcounter{figure}{0}

\begin{figure}[h]
    \centering
    \includegraphics[width=0.8\textwidth]{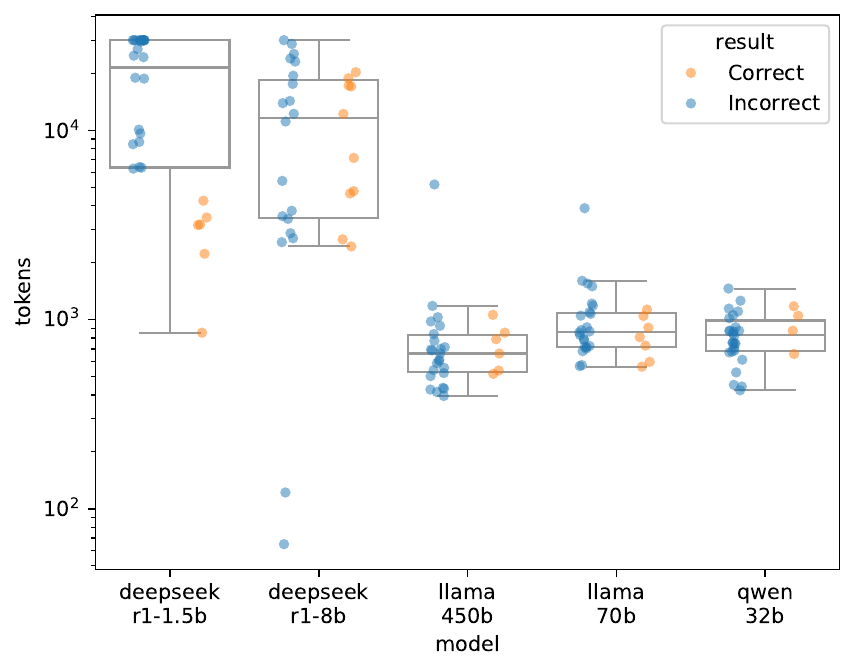}
    \caption{Boxplot of the generated token distribution for different models with reasoning (DeepSeek-R1-1.5B and 8B) and without reasoning (llama-3.3-70b-instruct, llama-3.1-405b-instruct, and qwen2.5-32b-instruct) in solving the AIME24 problem. Each dot represents an answer to a question for the corresponding model.}
    \label{sub_fig:boxplot_aime}
\end{figure}

\begin{figure}[h]
    \centering
    \includegraphics[width=0.8\textwidth]{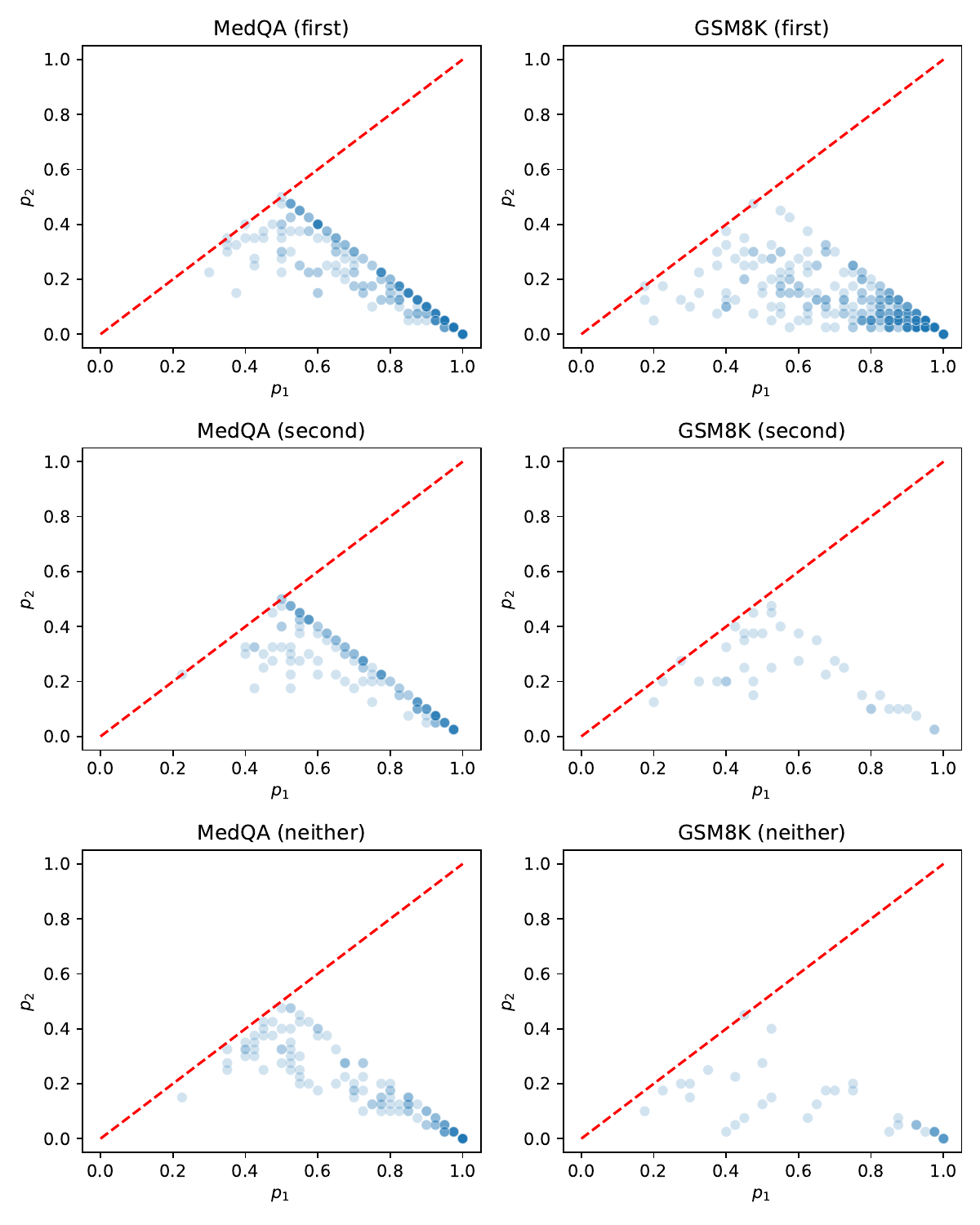}
    \caption{Scatterplot of the two most frequent responses for MedQA and GSM8K by OpenAI GPT-4o-mini with CoT. }
    \label{sub_fig:scatter}
\end{figure}

\begin{figure}[h]
    \centering
    \includegraphics[width=0.8\textwidth]{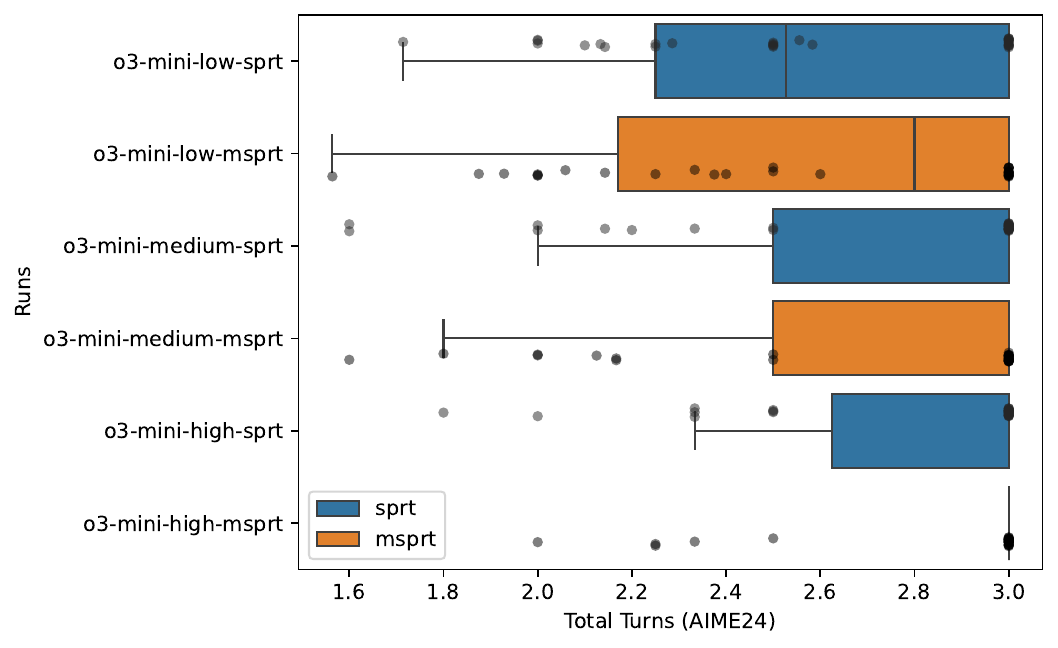}
    \caption{Boxplot of the average number of concurrently sampled LLM responses 
 by each turn for AIME24 benchmarks with our methods.}
    \label{sub_fig:figure-boxplot-for-concurrency-for-each-turn.pdf}
\end{figure}

\end{document}